\documentclass{article}

\usepackage{fullpage}
\usepackage[utf8]{inputenc}
\usepackage[T1]{fontenc}
\usepackage{microtype}
\usepackage{indentfirst}

\usepackage{amsmath}
\usepackage{amsfonts}
\usepackage{amssymb}
\usepackage{amsthm}
\usepackage{mathtools}
\usepackage{bm}
\usepackage{commath}
\usepackage{relsize}
\usepackage{bbm}
\usepackage{nicefrac}
\usepackage{mysymbol}

\usepackage{graphicx}
\usepackage{subfigure}
\usepackage{diagbox}
\usepackage{multirow}
\usepackage{booktabs}
\usepackage{hhline}
\usepackage{array}
\usepackage{pgfplots}
\usepackage{tikz}
\usetikzlibrary{positioning}

\usepackage{color}
\usepackage{tcolorbox}

\usepackage{algorithm}
\usepackage{algorithmic}
\usepackage{listings}

\usepackage{cite}
\usepackage{url}

\usepackage{pifont}
\usepackage{newfloat}
\usepackage{enumerate}
\usepackage{enumitem}
\usepackage[title]{appendix}
\usepackage[font={small}]{caption}

\newtheorem{theorem}{Theorem}[section]

\theoremstyle{definition}
\newtheorem{definition}[theorem]{Definition}
\newtheorem{assumption}[theorem]{Assumption}
\theoremstyle{remark}

\allowdisplaybreaks
\newcolumntype{?}{!{\vrule width 1pt}}

\makeatletter
\def\Let@{\def\\{\notag\math@cr}}
\makeatother

\usepackage[hidelinks,backref=page]{hyperref}
\definecolor{darkred}{RGB}{150,0,0}
\definecolor{darkgreen}{RGB}{0,150,0}
\definecolor{darkblue}{RGB}{0,0,150}
\hypersetup{colorlinks=true, linkcolor=darkred, citecolor=darkblue, urlcolor=darkblue}

\renewcommand*{\backref}[1]{}
\renewcommand*{\backrefalt}[4]{%
    \ifcase #1 (Not cited.)%
    \or        (Cited on page~#2.)%
    \else      (Cited on pages~#2.)%
    \fi}
    
\title{\textbf{Boosting Cross-problem Generalization in Diffusion-Based Neural Combinatorial Solver via Inference Time Adaptation}}

\date{}

\author{
Haoyu LEI\thanks{Department of Computer Science and Engineering, The Chinese University of Hong Kong, hylei22@cse.cuhk.edu.hk}, 
Kaiwen Zhou\thanks{Huawei Noah's Ark Lab, zhoukaiwen2@huawei.com}, 
Yinchuan Li\thanks{Huawei Noah's Ark Lab, liyinchuan@huawei.com},
Zhitang Chen\thanks{Huawei Noah's Ark Lab, chenzhitang2@huawei.com},
Farzan~Farnia\thanks{Department of Computer Science and Engineering, The Chinese University of Hong Kong, farnia@cse.cuhk.edu.hk}
	}

\begin{document}
\maketitle

\begin{abstract} 

Diffusion-based Neural Combinatorial Optimization (NCO) has demonstrated effectiveness in solving NP-complete (NPC) problems by learning discrete diffusion models for solution generation, eliminating hand-crafted domain knowledge. Despite their success, existing NCO methods face significant challenges in both cross-scale and cross-problem generalization, and high training costs compared to traditional solvers. While recent studies on diffusion models have introduced training-free guidance approaches that leverage pre-defined guidance functions for conditional generation, such methodologies have not been extensively explored in combinatorial optimization. To bridge this gap, we propose a training-free inference time adaptation framework (\textbf{DIFU-Ada}) that enables both the zero-shot cross-problem transfer and cross-scale generalization capabilities of diffusion-based NCO solvers without requiring additional training. We provide theoretical analysis that helps understanding the cross-problem transfer capability. Our experimental results demonstrate that a diffusion solver, trained exclusively on the Traveling Salesman Problem (TSP), can achieve competitive zero-shot transfer performance across different problem scales on TSP variants, such as Prize Collecting TSP (PCTSP) and the Orienteering Problem (OP), through inference time adaptation.

\end{abstract}

\section{Introduction}
Combinatorial optimization (CO) problems are fundamental challenges across numerous domains, from logistics and supply chain management to network design and resource allocation. While traditional exact solvers and heuristic methods have been widely studied, they often struggle with scalability and require significant domain expertise to design problem-specific algorithms \cite{arora1998polynomial,gonzalez2007handbook}.

Recent advances in deep learning have sparked interest in Neural Combinatorial Optimization (NCO), which aims to learn reusable solving strategies directly from data, eliminating the need for hand-crafted heuristics \cite{bengio2021machine}. Among various deep learning approaches, diffusion-based models \cite{ho2020denoising,song2020score} have emerged as a particularly promising direction for solving combinatorial optimization problems. These models have demonstrated remarkable capabilities in learning complex solution distributions by adapting discrete diffusion processes to graph structures \cite{sun2023difusco}. Recent works like \cite{li2024distribution,li2024fast} have achieved state-of-the-art performance on classical problems such as the Traveling Salesman Problem (TSP), showcasing the potential of diffusion-based generative approaches in combinatorial optimization.

However, the practical applicability of existing NCO approaches is limited by several generalization challenges. First, current models suffer from cross-scale generalization, with performance degrading significantly when applied to larger problem instances than those seen during training, especially for auto-regression-based solvers \cite{khalil2017learning,kool2018attention} including transformer and reinforcement learning methods. Second, these models show limited cross-problem transfer capabilities, struggling to adapt to problem variants with modified objectives or additional constraints. While several studies have attempted to enhance learning-based solvers' generalization through approaches such as training additional networks \cite{wang2024asp} and fine-tuning \cite{lin2024cross}, these methods require substantial computational costs and training data for training separate models for each problem type and scale.

\begin{figure*}[t]
    \centering
    \includegraphics[width=0.99\linewidth]{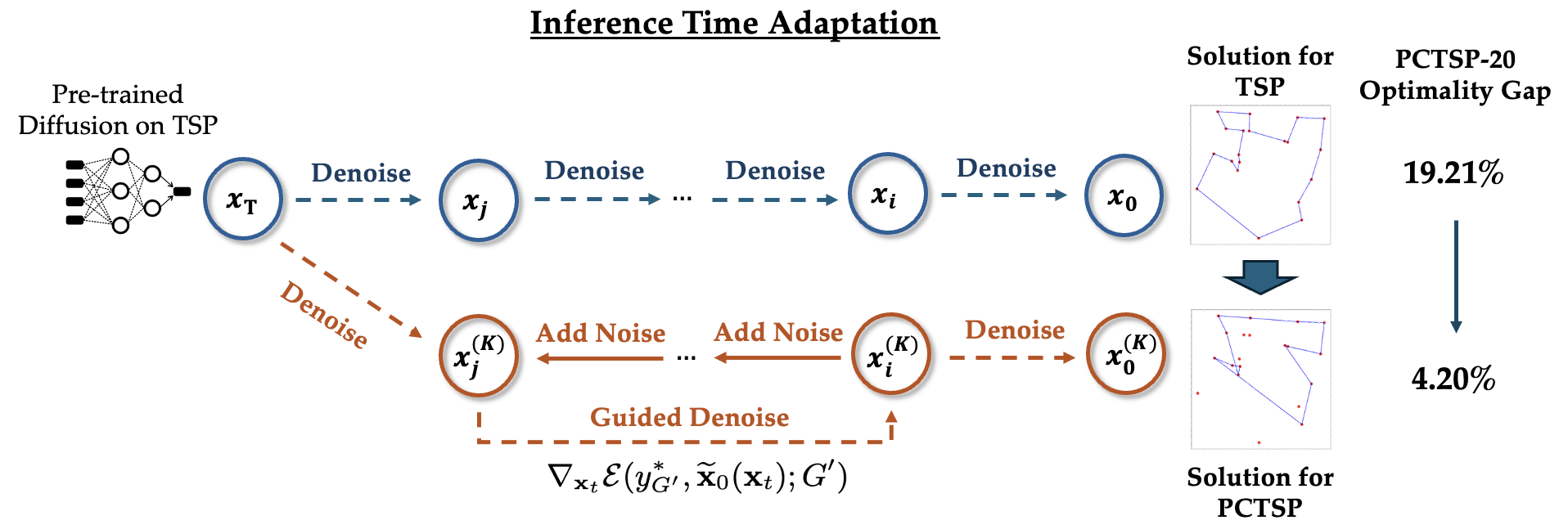}
    \caption{The proposed \textbf{Inference Time Adaptation} framework. This approach combines (1) energy-guided sampling, which incorporates problem-specific objectives and constraints, with (2) a recursive renoising-denosing travel for solution refinement, enabling zero-shot cross-problem transfer without training. The Optimality Gap ($\downarrow$) on PCTSP-20 is reduced from $19.21\%$ to $4.20\%$.}
    \label{img:framework}
\end{figure*}

In parallel, recent advances in diffusion models, particularly in computer vision, have demonstrated the effectiveness of training-free guidance approaches for enhancing conditional generation \cite{bansal2023universal,chung2022diffusion,yu2023freedom,shen2024understanding,ye2024tfg}. These approaches leverage plug-and-play guidance functions or pre-trained networks to enable conditional generation without additional training overhead. These approaches demonstrate significant potential for modifying the inference phase to achieve controllable sampling from pre-trained diffusion models. Inspired by these developments, we explore the training-free inference time adaptation to address the cross-problem transfer challenges in neural combinatorial optimization in Figure~\ref{img:framework}.

This work introduces an inference time adaptation framework (illustrated in Figure~\ref{img:framework}) designed to enhance the generalization capabilities of diffusion-based Neural Combinatorial Optimization (NCO) solvers without additional training costs. We scale up the inference time by combining two key components: (1) energy-guided sampling and (2) a recursive renoising-denosing travel modelled as a Guided Langevin Dynamics. Our DIFU-Ada framework enables zero-shot cross-problem transfer while maintaining solution feasibility. Through experimental evaluation of the TSP-trained diffusion model on more complex variants, the Prize Collecting TSP (PCTSP) and the Orienteering Problem (OP), we empirically demonstrate its effective zero-shot transferability across different problems of increasing complexity, while maintaining consistent performance across different problem scales. Our work represents a significant step toward more flexible and generalizable diffusion-based combinatorial optimization solvers, reducing the need for problem-specific model training while maintaining competitive performance.

\section{Related Works}
\noindent\textbf{Neural Network-based Combinatorial Solvers.}
Neural Combinatorial Optimization (NCO) leverages neural networks to learn solution distributions for complex optimization problems \cite{bengio2021machine, zhang2023survey}. Prominent approaches include autoregressive solvers \cite{khalil2017learning, kool2018attention, kwon2020pomo, kim2022sym, hottung2021learning} and non-autoregressive methods \cite{joshi2019efficient, fu2021generalize, qiu2022dimes, wang2024asp, sun2023difusco, sanokowski2024diffusion}.

\noindent\textbf{Diffusion-based Generative Modeling.}
Score-based diffusion models \cite{ho2020denoising, song2020score, sohl2015deep, song2019generative, dhariwal2021diffusion, song2020denoising} have become a leading generative framework. Their application to CO was pioneered by \cite{sun2023difusco} for the Traveling Salesman Problem and has been rapidly advanced by methods incorporating gradient search \cite{li2024distribution}, single-step consistency models \cite{li2024fast}, and unsupervised frameworks \cite{sanokowski2024diffusion}.

\noindent\textbf{Training-free Guidance for Diffusion Models.}
To control generation, training-free guidance methods \cite{bansal2023universal, chung2022diffusion, yu2023freedom, shen2024understanding，zhou2026learning} offer an efficient alternative to training-intensive classifier \cite{dhariwal2021diffusion} or classifier-free \cite{ho2022classifier} approaches. This strategy was first adapted for discrete diffusion solvers by \cite{li2024distribution}, building on the framework from \cite{sun2023difusco}. Our work further demonstrates that modifying the inference phase is a powerful method for enhancing the zero-shot cross-problem transfer of diffusion-based NCO solvers.

\section{Preliminaries}
\subsection{Graph-based CO Problems}
Combinatorial optimization (CO) problems on graphs are fundamental to numerous real-world applications. Following recent advances \cite{sun2023difusco,li2024distribution}, we address these problems by formalizing graph-based CO instances. We represent each problem instance as an undirected graph $G(V,E)\in\mathcal{G}$, where $V$ and $E$ denote the vertex and edge sets, respectively. This representation encompasses both vertex selection and edge selection problems, covering a broad spectrum of practical CO scenarios. For any instance $G\in\mathcal{G}$, we define a binary decision variable $\mathbf{x}\in\mathcal{X}_{\mathcal{G}}$, where $\mathcal{X}_{\mathcal{G}}=\{0,1\}^{N}$ represents the feasible solution space. The optimization objective is to find the optimal solution $\mathbf{x}^*$ that minimizes a problem-specific objective function $\phi(\cdot;G):\{0,1\}^{N}\rightarrow\mathbb{R}$:
\begin{align}
\mathbf{x}^* = \underset{\mathbf{x}\in\mathcal{X}_{\mathcal{G}}}{\text{argmin}}\; \phi(\mathbf{x}; G),
\end{align}
where the objective function decomposes into a log-barrier formulation \cite{den1992classical}:
\begin{align}
\phi(\mathbf{x}; G) = f_{\text{cost}}(\mathbf{x}; G) - \mu \sum_{i=1}^m \log\big(-g_i(\mathbf{x}; G)\big),
\end{align}
Here $f_{\text{cost}}(\cdot;G)$ measures the solution quality, and $g_i(\cdot;G)$ enforces problem-specific constraints through a penalty coefficient $\mu>0$. The validity functions $g_i(\cdot;G)$ returns $0$ for feasible solutions and is strictly positive for \mbox{infeasible ones}. 

\subsection{Energy-based Probabilistic Modeling}
To leverage recent advances in deep generative models, we reformulate the CO objective through an energy-based perspective \cite{lucas2014ising}. Specifically, we establish an energy function:
\begin{align}
    \mathcal{E}(\cdot;G):=\rho(y,\phi(\cdot;G)), 
\end{align}
where $\rho$ represents any loss functions (cross entropy loss, L2-norm), and the energy function maps each solution to its corresponding energy state. This energy-based formulation naturally leads to a probabilistic framework through the Boltzmann distribution \cite{lecun2006tutorial}:
\begin{align}
&p(y|\mathbf{x};G) = \frac{\exp\big(-\frac{1}{\tau}\mathcal{E}(y,\mathbf{x};G)\big)}{\mathcal{Z}}, \\
\text{where}\;&\mathcal{Z} = \sum_{\mathbf{x}}\exp\Big(-\frac{1}{\tau}\mathcal{E}(y,\mathbf{x};G)\Big),
\label{eq:boltzmann}
\end{align}
where $\tau$ controls the temperature of the system and $\mathcal{Z}$ denotes the partition function that normalizes the distribution. Recent works have demonstrated promising approaches to approximate this distribution using diffusion-based deep generative models by parameterizing a conditional distribution $p_{\theta}(\mathbf{x}|G)$ to minimize the energy function. Both supervised \cite{sun2023difusco,li2024distribution,li2024fast} and unsupervised \cite{sanokowski2024diffusion} learning paradigms have shown significant advances. Since our proposed training-free guidance mechanism is applicable to any pre-trained diffusion-based solver, we focus on the supervised learning framework in this work.

\subsection{Discrete Diffusion Generative Models}
Given a training set $\mathcal{G}=\{G_i\}_{i=1}^k$ of i.i.d. problem instances with their optimal solutions $\mathbf{x}$ and the corresponding optimal objective values $y_G^*$, we use generative model and optimize the model parameters $\theta$ by maximizing the likelihood of the optimal solutions:
\begin{equation}
L(\theta) = \mathbb{E}_{G\sim\mathcal{G}}[-\log p_{\theta}(\mathbf{x}|y_G^*,G)].
\end{equation}
We adopt a discrete diffusion generative model \cite{austin2021structured} to effectively sample optimal solutions from the learned distribution $p_{\theta}(\mathbf{x}|y^*,G)$~\cite{sun2023difusco, li2024distribution, li2024fast}.

The diffusion process consists of two key components: a forward process that gradually corrupts the data, and a reverse process that learns to reconstruct the original distribution. The forward process $q(\mathbf{x}_{1:T}|\mathbf{x}_{0})=\prod_{t=1}^{T}q(\mathbf{x}_{t}|\mathbf{x}_{t-1})$ maps clean data $\mathbf{x}_{0}\sim q(\mathbf{x}_{0}|G)$ to a sequence of increasingly corrupted latent variables $\mathbf{x}_{1:T}$. The reverse process $p_{\theta}(\mathbf{x}_{0:T}|G)=p(\mathbf{x}_{T})\prod_{t=1}^{T}p_{\theta}(\mathbf{x}_{t-1}|\mathbf{x}_{t},G)$ learns to gradually denoise these latent variables to recover the original distribution. For discrete state spaces, we define the forward process using a categorical distribution:
\begin{align}
q(\mathbf{x}_{t}|\mathbf{x}_{t-1})= \text{Cat}(\mathbf{x}_{t};\mathbf{p}=\mathbf{\widetilde{x}}_{t-1}\mathbf{Q}_{t}),
\end{align}
where $\mathbf{\widetilde{x}}_{t}\in\{0,1\}^{N\times2}$ represents the one-hot encoding of $\mathbf{x}_{t}\in\{0,1\}^{N}$. The forward transition matrix $\mathbf{Q}_t$ is defined as:
\begin{align}
\mathbf{Q}_t = \begin{bmatrix} (1-\beta_t) & \beta_t \\ \beta_t & (1-\beta_t) \end{bmatrix}, \quad \beta_t\in[0,1],
\end{align}
where $[\mathbf{Q}_t]_{ij}$ denotes the state transition probability from state $i$ to state $j$. The $t$-step marginal distribution and posterior can be derived as:
\begin{equation}
\begin{aligned}
q(\mathbf{x}_t|\mathbf{x}_0) &= \text{Cat}(\mathbf{x}_t; \mathbf{p} = \mathbf{\widetilde{x}}_0\overline{\mathbf{Q}}_t), \\
\end{aligned}
\end{equation}
where $\overline{\mathbf{Q}}_t = \mathbf{Q}_1\mathbf{Q}_2\ldots\mathbf{Q}_t$ and $\odot$ denotes element-wise multiplication. 

To capture the structural properties of CO problems, we employ an anisotropic graph neural network architecture \cite{joshi2019efficient}. For a given instance $G$, the network learns to predict the clean data distribution $p_{\theta}(\mathbf{\widetilde{x}}_0|\mathbf{x}_t,G)$. Taking TSP as an example, where $G$ encodes the 2D Euclidean coordinates of vertices, the network outputs an adjacency matrix $p_{\theta}(\mathbf{\widetilde{x}}_0|\mathbf{x}_t,G)\in [0,1]^{N\times 2}$. This matrix parameterizes $N$ independent Bernoulli distributions, each corresponding to a binary decision variable in $\mathbf{\widetilde{x}}_0$. The reverse process during sampling follows:
\begin{align}
p_\theta(\mathbf{x}_{t-1}|\mathbf{x}_t,G) = \sum_{\widetilde{\mathbf{x}}_0} q(\mathbf{x}_{t-1}|\mathbf{x}_t, \widetilde{\mathbf{x}}_0)p_\theta(\widetilde{\mathbf{x}}_0|\mathbf{x}_t,G).
\label{reverse}
\end{align}


\section{Method: Inference Time Adaptation}
\subsection{Energy-guided Sampling for Problem Transfer}
While training-free guidance has been extensively studied in computer vision \cite{bansal2023universal, chung2022diffusion, yu2023freedom, shen2024understanding,ye2024tfg}, its application to combinatorial optimization problems has only recently been explored by \cite{li2024distribution, li2024fast}. Adopted by \cite{li2024distribution, li2024fast}, we extend this approach by introducing energy-based training-free guidance for zero-shot cross-problem transfer during sampling, enabling flexible incorporation of additional problem-specific objective functions and constraints into pre-trained diffusion-based CO solvers.


Let $\mathcal{G}'=\{G'_i\}_{i=1}^n$ denote a set of new problem instances which differs from the pre-trained problem set $\mathcal{G}$. For a new instance $G'$ with its optimal solution pair $(\mathbf{x}, y_{G'}^*)$, we need to estimate the new reverse process $p_\theta(\mathbf{x}_{t-1}|\mathbf{x}_t,y_{G'}^*,G')$ according to (\ref{reverse}). Following the score estimation perspective of diffusion processes \cite{song2020score, dhariwal2021diffusion}, the reverse sampling for new problem types $G'$ can be defined in a SDE formulation:
\begin{equation}
\label{eq:energy_guided_sde}
d\mathbf{x} = \left[ -\mathbf{f}(\mathbf{x}, t) + g^2(t) \hat{s}_\theta(\mathbf{x}, t, G') \right] dt' + g(t)d\mathbf{w}
\end{equation}
where $t = T - t'$. We decompose the conditional score function $\hat{s}_\theta(\mathbf{x}, t, G')$ at time step $t$ into two components:
\begin{equation}
\begin{aligned}
\hat{s}_\theta(\mathbf{x}, t, G') :=\underbrace{\nabla_{\mathbf{x}_t} \log p_\theta(\mathbf{x}|y^*_{G'},G')}_{\text{Posterior Score}} = \underbrace{\nabla_{\mathbf{x}_t} \log p_\theta(\mathbf{x}_t|G')}_{\text{Pre-trained Prior Score}} + \underbrace{\nabla_{\mathbf{x}_t} \log p_t(y^*_{G'}|\mathbf{x}_t,G')}_{\text{Energy Potential}}.
\end{aligned}
\label{eq:log_decomp}
\end{equation}
From the Bayesian perspective, $p_\theta(\mathbf{x}_t|G')$ can be understood as the \textit{prior}, which contains knowledge of the pre-trained problems (i.e. TSP problem in our experimental settings), and $p_t(y^*_{G'}|\mathbf{x}_t,G')$ corresponds to the \textit{Energy Potential} that incorporates additional constraints or objectives of the variant problem. We sample from the \textit{Posterior Score} $\hat{s}_\theta(\mathbf{x}, t, G')$ to generate high-quality solutions to the new problem $G'$. A theoretical analysis of how the information contained in pre-trained models can benefit the guided sampling is provided in Appendix~\ref{Appendix:analysis}. 

Drawing upon this theoretical framework, we leverage the pre-trained diffusion model to estimate the first term $\nabla_{\mathbf{x}} \log p_\theta(\mathbf{x}_t|G')$. In the context of cross-problem transfer, while the pre-trained model yields only a \textit{biased} score function for new problem instances, we compute the second energy-guided term $\nabla_{\mathbf{x}_t} \log p_t(y^*_{G'}|\mathbf{x}_t,G')$ to adjust using an energy function that specifically accounts for the additional objectives and constraints of the new problems:
\begin{equation}
\begin{aligned}
    \nabla_{\mathbf{x}_t} \log p_t(y^*_{G'}|\mathbf{x}_t,G') &\propto -\nabla_{\mathbf{x}_t}\mathcal{E}(y^*_{G'}, \widetilde{\mathbf{x}}_0(\mathbf{x}_t);G'),
\label{eq:energy_grad}
\end{aligned}
\end{equation}
where $\rho(y^*_{G'}, \phi( \widetilde{\mathbf{x}}_0(\mathbf{x}_t);G'))$ measures the distance of energy states between the optimal value and the predicted solution. Here, $\widetilde{\mathbf{x}}_0(\mathbf{x}_t)$ represents the predicted clean sample from the current noisy state $\mathbf{x}_t$, which is estimated by the logits of $N$ independent Bernoulli samples $\widetilde{\mathbf{x}}_0(\mathbf{x}_t)=\mathbb{E}_{\widetilde{\mathbf{x}}_0\sim p_\theta(\widetilde{\mathbf{x}}_0|\mathbf{x}_t)}[\widetilde{\mathbf{x}}_0]$. In this work, we define $\rho(y^*_{G'}, \phi(\widetilde{\mathbf{x}}_0(\mathbf{x}_t); G')) := \| y^*_{G'} - \phi(\widetilde{\mathbf{x}}_0(\mathbf{x}_t); G') \|_2$. From Eq~\eqref{eq:energy_grad}, we have:
\begin{equation}
\begin{aligned}
    \nabla_{\mathbf{x}_t} \log p_t(y^*_{G'}|\mathbf{x}_t,G') &\propto -\nabla_{\mathbf{x}_t}\rho(y^*_{G'}, \phi( \widetilde{\mathbf{x}}_0(\mathbf{x}_t);G')), \\ &\propto -\nabla_{\mathbf{x}_t} \phi(\widetilde{\mathbf{x}}_0(\mathbf{x}_t); G'),
\end{aligned}
\label{Eq:guided_score}
\end{equation}



Now we obtain the \textit{Energy Potential} by just taking the gradient of the problem-specific objective function $\phi(\widetilde{\mathbf{x}}_0; G)$. Given a pre-trained score-based diffusion model $s_\theta(\mathbf{x}, t, G) \approx \nabla_{\mathbf{x}_t} \log p_\theta(\mathbf{x}_t|G')$,the reverse process is guided by \textit{Energy Potential} modifying the score estimate as follows:
\begin{equation}
\label{eq:energy_guided_sde}
\begin{aligned}
d\mathbf{x} = \bigg[ &-\mathbf{f}(\mathbf{x}, t) + g(t)^2 \bigg( \underbrace{\nabla_{\mathbf{x}} \log p_\theta(\mathbf{x}|G')}_{\text{Pre-trained Score}} - \underbrace{\tau \nabla_{\mathbf{x}} \phi(\widetilde{\mathbf{x}}_0(\mathbf{x}); G')}_{\text{Energy Potential}} \bigg) \bigg] dt’ + g(t)d\mathbf{w}
\end{aligned}
\end{equation}
where $t=T-t'$. For discrete diffusion, each step of the reverse process involves sampling the next state, $\mathbf{x}_{t}$, from a multivariate Bernoulli distribution. This distribution is parameterized by the output of the guided posterior $p_\theta(\mathbf{x}| y^*_{G'},G')$, following a standard sampling framework like DDIM~\cite{song2020denoising}.

\begin{figure}[t]
    \centering
    \includegraphics[width=0.66\linewidth]{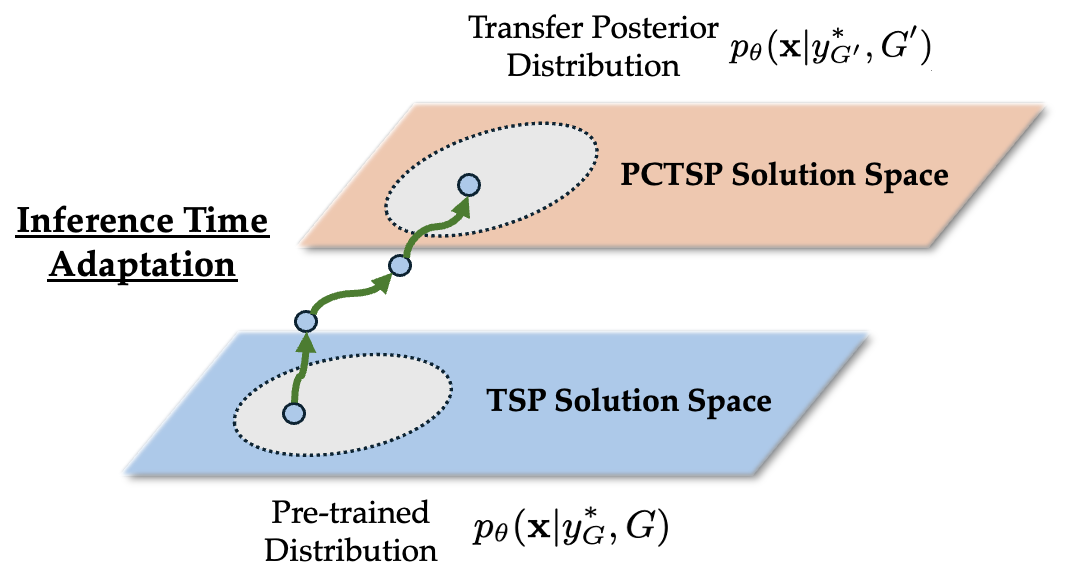}
    \caption{Overview of recursive renoising-denoising travel in \textbf{Inference Time Adaptation} for achieving zero-shot cross-problem generalization, sequentially shifting from pre-trained problem $G$ (TSP) solution distribution to target problem distribution $G'$ (PCTSP).}
    \label{img:overview}
\end{figure}

In this work, we have the well-established log-barrier formulations for PCTSP (Definition~\ref{eq.pctsp}) and OP (Definition~\ref{eq.op}) as the plug-and-play \textit{Energy Potential} following \cite{lucas2014ising}: 
\begin{definition}[Energy Potential for Prize Collecting TSP (PCTSP)] \label{eq.pctsp}
Given a complete graph $G=(V,E)$ with edge weights $w:E\rightarrow\mathbb{R}^+$, vertex prizes $r:V\rightarrow\mathbb{R}^+$, penalties $p:V\rightarrow\mathbb{R}^+$, and prize threshold $R$, find $\mathbf{x}\in\{0,1\}^{|E|}$, $\mathbf{y}\in\{0,1\}^{|V|}$ that minimizes:
\begin{align*}
\phi(\mathbf{x}, \mathbf{y};G) &= f_{\text{cost}}(\mathbf{x}, \mathbf{y};G) - \mu\cdot \log(-g(\mathbf{x}, \mathbf{y};G)) \\
\text{where}\quad f_{\text{cost}}(\mathbf{x},\mathbf{y};G) &= \sum_{e\in E} w_e x_e + \sum_{v\in V} p_v(1-y_v) \\
g(\mathbf{x},\mathbf{y};G) &= \max(0, R-\sum_{v\in V} r_v y_v)
\end{align*}
\end{definition}

\begin{definition}[Energy Potential for Orienteering Problem (OP)] \label{eq.op}
Given a complete graph $G=(V,E)$ with edge weights $w:E\rightarrow\mathbb{R}^+$, vertex scores $s:V\rightarrow\mathbb{R}^+$, and budget $B$, find $\mathbf{x}\in\{0,1\}^{|E|}$, $\mathbf{y}\in\{0,1\}^{|V|}$ that minimizes:
\begin{align*}
\phi(\mathbf{x}, \mathbf{y};G) &= f_{\text{cost}}(\mathbf{x}, \mathbf{y};G) - \mu\cdot \log(-g(\mathbf{x}, \mathbf{y};G)) \\
\text{where}\quad f_{\text{cost}}(\mathbf{x},\mathbf{y};G) &= -\sum_{v\in V} s_v y_v \\
g(\mathbf{x},\mathbf{y};G) &= \max(0, \sum_{e\in E} w_e x_e - B)
\end{align*}
\end{definition}
\subsection{Decoding from Generated Heatmaps}
Distinguished from auto-regressive solvers like transformer, diffusion-based solvers for combinatorial optimization learn to generate a time-dependent matrix of probabilities, often visualized as an $N\times N$ adjacency matrix (heatmap). This heatmap represents the likelihood of each element (e.g., an edge in a graph) being part of the final solution, rather than the discrete solution itself. Consequently, a decoding strategy is essential to translate this probabilistic output into a feasible, binary solution structure.

A common and direct approach is greedy decoding, which iteratively constructs a solution by selecting the elements with the highest probabilities from the heatmap until a valid termination condition is met (e.g., forming a complete tour in the Traveling Salesperson Problem). Alternatively, the generated heatmaps can be post-processed by more sophisticated improvement heuristics, such as 2-opt~\cite{lin1973efficient, li2024distribution} or Monte Carlo Tree Search~\cite{sun2023difusco}. To ensure a fair comparison across all methods in our experiments, we apply greedy decoding.

\subsection{Recursive Renoising-denoising Travel}
Our preliminary experiments revealed that applying energy-guided sampling alone is often insufficient for zero-shot, cross-problem transfer. The distributional divergence between the source problem ($G$) and the target problem ($G'$) can hinder the generation of high-quality, feasible solutions. To bridge this gap, we propose 2-phase Inference-Time Adaptation framework combining energy-guided sampling and recursive renoising-denoising travel, inspired by recent work~\cite{song2023consistency, li2024distribution}. 

We theoretically interpret our Inference-time Adaptation as a Guided Langevin Dynamics~\cite{song2019generative, welling2011bayesian} process in Figure~\ref{img:overview}. It is designed to iteratively transport the solution particle from a target distribution $G$ towards a new problem distribution $G'$. Given a solution $x^{k}$ in iteration $k$, the recursive update to next iteration $k+1$ is a discrete-time sequence of SDE:
\begin{equation}
\label{eq:guided_langevin_update}
\mathbf{x}^{(k+1)} \leftarrow \mathbf{x}^{(k)} -\left( s_\theta(\mathbf{x}^{(k)}, G) - \tau \nabla_{\mathbf{x}^{(k)}}\phi(\mathbf{x}^{(k)};G')\right) +\sigma_k \mathbf{z}, 
\end{equation}
where $\mathbf{z}$ represent renoising. A naive implementation approach would be to recursively simulate the full SDE (i.e., repeated full re-noising and denoising) to progressively adapt the solution to the target distribution. However, this incurs a prohibitive computational cost. 

Our key insight is that the energy guidance does not need to be fully applied at every step. Based on this, we implement the recursive process to only a few renoising steps and a single step denoising in each iteration, achieving a 5-10x inference speedup compared to the full recursive approach.

Algorithm~\ref{alg:main} outlines the framework for Inference-Time Adaptation, combining energy-guided sampling and recursive renoising-denoising travel. It provides a practical and efficient method for zero-shot cross-problem transfer with pre-trained diffusion models at no additional training cost. Furthermore, our framework is model-free and can be augmented with other heuristics, such as 2-opt~\cite{lin1973efficient} or MCTS~\cite{fu2021generalize, sun2023difusco}, to further enhance final solution quality.

\begin{algorithm}[t]
\caption{Inference Time Adaptation for Cross-problem}
\label{alg:main}
\begin{algorithmic}[1]
\REQUIRE 
\STATE $p_\theta$: Pre-trained diffusion solver
\STATE $G$: Original problem
\STATE $G'$: Target problem
\STATE $T$: Number of diffusion steps
\STATE $K$: Number of recursive travel
\STATE $i$: Number of renoising steps
\STATE $\tau$: Energy guidance temperature
\ENSURE Optimal solution $y^{(K)}$ for problem instance $G'$

\STATE Initialize $\mathbf{x}^{(0)}_T\in\{0,1\}^N$ 
\FOR{$k = 1$ \TO $K$}
    \STATE Re-noise solution $\mathbf{x}_0^{(k-1)}$ to some noise level $\mathbf{x}_i^{(k)}$
    \FOR{$t = i$ \TO 0}
        \STATE Compute $p_\theta(\mathbf{x}_t^{(k)}|G')$ from pre-trained model
        \STATE Compute energy potential: $\nabla_{\mathbf{x}_t}\phi(\widetilde{\mathbf{x}}_0(\mathbf{x}^{(k)}_t);G')$
        \STATE Compute posterior $p_\theta(\mathbf{x}^{(k)}_{t-1}|\mathbf{x}^{(k)}_t, y^*_{G'},G')$
        \STATE Update next state with Bernoulli Sampling: $\mathbf{x}^{(k)}_{t-1} \sim \text{Cat}(\mathbf{x}^{(k)}_{t-1}; p_\theta(\mathbf{x}^{(k)}_{t-1}|\mathbf{x}^{(k)}_t, y^*_{G'},G'))$
    \ENDFOR
    \STATE Decode $\mathbf{x}^{(k)}_0$ to the optimal solution $y^{(k)}$
\ENDFOR
\RETURN $y^{(K)}$
\end{algorithmic}
\end{algorithm}

\section{Numerical Results}
\noindent\textbf{Dataset.} We evaluate our approach on the commonly-used NP-complete combinatorial optimization problems: the Traveling Salesman Problem (TSP) together with its variants, the Prize Collecting Traveling Salesman Problem (PCTSP) and Orienteering Problem (OP). The detailed problem descriptions are given in Appendix~\ref{app:def}. 

\noindent\textbf{Evaluation.} Following \cite{kool2018attention}, we generate 10000 test instances for each problem scale: 20, 50, and 100, which denotes the node counts for PCTSP and OP. We evaluate model performance using two primary metrics: average solution cost and optimality gap relative to the exact solution solvers. Additionally, we measure computational efficiency through total training time and per-instance inference time. 

\noindent\textbf{Baselines.} We compare our approach against multiple baseline categories. For PCTSP, (1) Exact solver: Gurobi; (2) OR-based heuristics: OR-Tools and Iterated Local Search (ILS); (3) Learning-based methods: AM \cite{kool2018attention}, MDAM \cite{xin2021multi}, AM-FT \cite{lin2024cross}, ASP \cite{wang2024asp}, DIFUSCO \cite{sun2023difusco}, T2T \cite{li2024distribution}. For OP, (1) Exact solver: Gurobi; (2) OR-based heuristics: Compass \cite{kobeaga2018efficient} and Tsili \cite{tsiligirides1984heuristic}; (3) Learning-based methods: AM \cite{kool2018attention}, AM-FT \cite{lin2024cross}, DIFUSCO \cite{sun2023difusco}, T2T \cite{li2024distribution}. 

\noindent\textbf{Experimental Setup.} The proposed Inference Time Adaptation for Diffusion-based solvers (\textbf{DIFU-Ada}) builds upon DIFUSCO's \cite{sun2023difusco} TSP-trained checkpoints in three different scales without additional training. The backbone archetechture used in diffusion model is the same as DIFUSCO, a 12-layer Anisotropic GNN with a width of 256 hidden nodes. The results of DIFU-Ada are recorded under 100 iterations, with 5-step re-noising and 1-step guided denoising. The decoding strategy used in experiments is greedy decoding. The energy functions used in guided sampling are stated in Proposition~\ref{eq.pctsp} and \ref{eq.op}. The guidance temperature $\lambda$ is fixed to be 0.1 and the constraint coefficient $\mu$ is set to be 1 for all problem instances. All experiments are conducted on a single Tesla V100 GPU. Experimental setup details are provided in Appendix~\ref{setup}.

\begin{table}[htbp]\LARGE
\renewcommand{\arraystretch}{1.5}
\centering
\resizebox{0.95\textwidth}{!}{
\begin{tabular}{llccccccccc}
\hline
\hline
& \multicolumn{3}{c}{\textbf{PCTSP-20}} & \multicolumn{3}{c}{\textbf{PCTSP-50}} & \multicolumn{3}{c}{\textbf{PCTSP-100}} \\
\cmidrule(l){2-4}
\cmidrule(l){5-7}
\cmidrule(l){8-10}
Method & \textbf{Cost}$\downarrow$ & \textbf{Gap}$\downarrow$ & \textbf{Time}$\downarrow$ & \textbf{Cost}$\downarrow$ & \textbf{Gap}$\downarrow$ & \textbf{Time}$\downarrow$ & \textbf{Cost}$\downarrow$ & \textbf{Gap}$\downarrow$ & \textbf{Time}$\downarrow$ \\
\hline
DIFUSCO & 3.78 & 19.21\% & 1.04s & 5.20 & 15.97\% & 1.35s & 7.85 & 32.31\% & 2.02s \\
T2T & 3.64 & 14.82\% & 2.02s & 4.90 & 9.38\% & 2.18s & 7.23 & 21.92\% & 2.70s \\

\textbf{DIFU-Ada (Ours)} & \textbf{3.30} & \textbf{4.20\%} & 1.72s  & \textbf{4.63} & \textbf{3.57\%} & 1.99s & \textbf{6.51} & \textbf{9.61\%} & 2.74s \\
\hline
\end{tabular}
}
\\ 
\resizebox{0.95\textwidth}{!}{
\begin{tabular}{llccccccccc}
\hline
& \multicolumn{3}{c}{\textbf{OP-20}} & \multicolumn{3}{c}{\textbf{OP-50}} & \multicolumn{3}{c}{\textbf{OP-100}} \\
\cmidrule(l){2-4}
\cmidrule(l){5-7}
\cmidrule(l){8-10}
Method & \textbf{Prize}$\uparrow$ & \textbf{Gap}$\downarrow$ & \textbf{Time}$\downarrow$ & \textbf{Prize}$\uparrow$ & \textbf{Gap}$\downarrow$ & \textbf{Time}$\downarrow$ & \textbf{Prize}$\uparrow$ & \textbf{Gap}$\downarrow$ & \textbf{Time}$\downarrow$ \\
\hline
DIFUSCO & 9.25 & 12.48\% & 1.51s & 25.60 & 13.45\% & 1.88s & 45.66 & 20.02\% & 2.04s \\
T2T & 9.67 & 8.51\% & 2.70s & 26.89 & 9.09\% & 3.04s & 48.72 & 14.70\% & 3.45s \\

\textbf{DIFU-Ada (Ours)} & \textbf{10.24} & \textbf{3.11\%} & 2.09s & \textbf{28.21} & \textbf{4.63\%} & 2.35s & \textbf{54.56} & \textbf{8.06\%} & 2.91s \\
\hline
\hline
\end{tabular}
}
\caption{Zero-shot cross-problem transfer performance comparison between TSP-trained DIFUSCO, T2T and DIFU-Ada (Ours) approach on PCTSP and OP instances. The results show solution cost, optimality gap, and the average inference time for one instance across three problem scales (20, 50, and 100 nodes).}
\label{tab:op_abla}
\end{table}

\subsection{Cross-problem and Cross-scale Generalization}

To validate the effectiveness of our inference time adaptation framework in zero-shot cross-problem transfer scenarios, we first conduct experiments while maintaining consistent problem scales. As demonstrated in Table \ref{tab:op_abla}, our method achieves significant zero-shot performance improvements when applied to the pre-trained DIFUSCO \cite{sun2023difusco} and T2T \cite{li2024distribution} model on TSP. These improvements are evident across all PCTSP and OP problem scales, in both enhanced solution quality and reduced optimality gaps compared to the exact OR solvers.

Following the evaluation strategy following \cite{wang2024asp}, we conducted experiments of DIFU-Ada's transferability across diverse combinatorial optimization (CO) problem types and scales. Our evaluation framework specifically assesses two aspects of generalization: 1) \textbf{Cross-problem Generalization}: We assess the model's adaptability to different CO problem variants, such as transferring from TSP to PCTSP or OP. 2) \textbf{Cross-scale Generalization}: We evaluate the performance of a single, optimally trained model across various problem scales (e.g., from small to large instances of the same problem type). Unlike existing learning-based methods that necessitate problem-specific retraining for new problem types or scales (e.g., for PCTSP and OP instances), our DIFU-Ada framework achieves zero-shot generation at inference time. This rigorous out-of-distribution evaluation strategy presents a more challenging benchmark than conventional in-distribution assessments, testing the model's capacity of generalization.

\begin{figure}[htbp] 
\centering    
\includegraphics[width=0.9\linewidth]{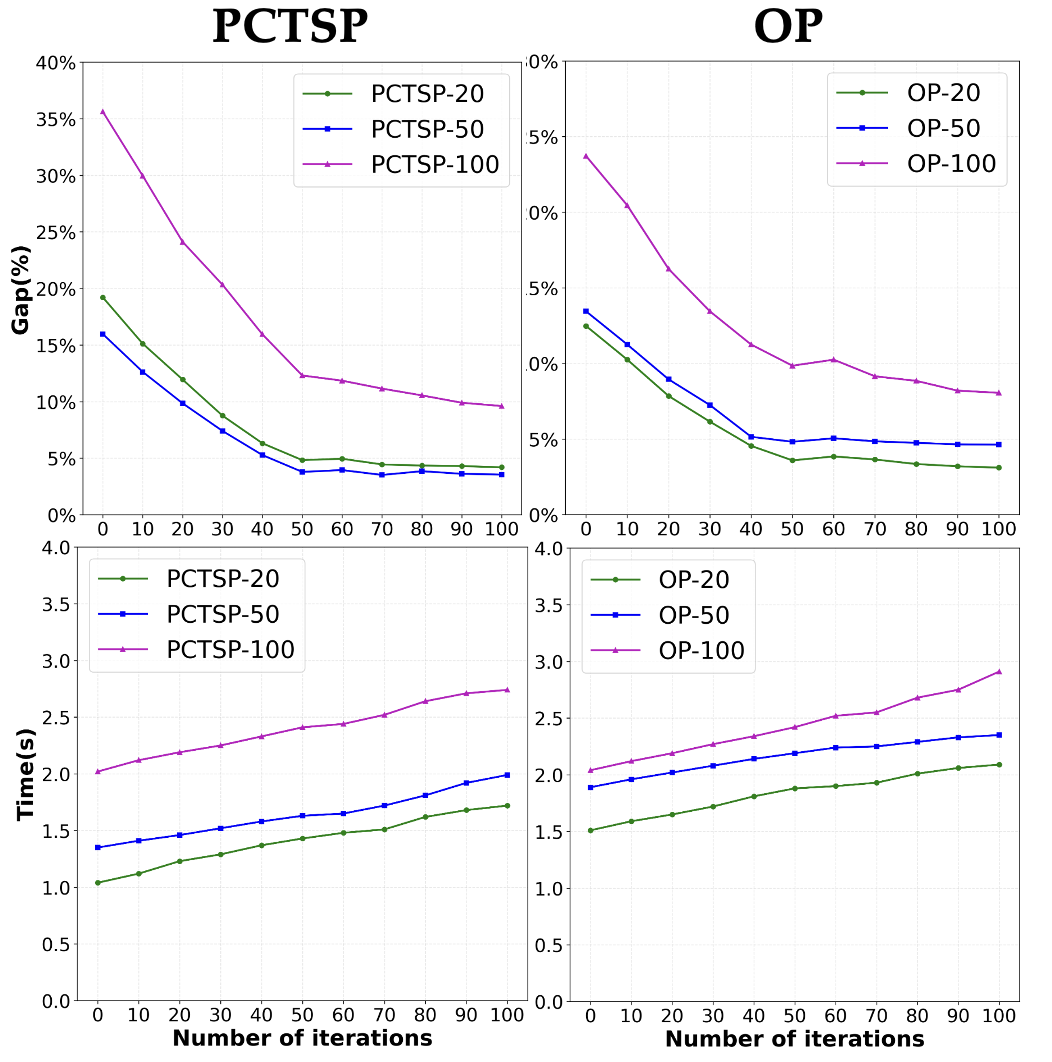}
\caption{Ablation studies of the number of recursive travel steps on the trade-off between optimality gap (\%) and inference time (s) for PCTSP and OP.}
\label{Fig: tradeoff}
\end{figure}

The experimental results presented in Table \ref{tab:pctsp} (PCTSP) and Table \ref{tab:op} (OP) show the cross-problem and cross-scale generalization capabilities of DIFU-Ada. Our framework achieves competitive average optimality gaps ($9.70\%$ for PCTSP and $7.26\%$ for OP) across all scales, while maintaining its zero-shot cross-problem transfer capability. This performance is achieved without the training costs (e.g., 3-5 days of retraining reported for existing methods) or the need for well-labeled training datasets. While OR-based heuristics like Compass exhibit strong performance on narrowly defined problem types, DIFU-Ada's ability to leverage pre-trained diffusion models across different CO problem variants without additional training highlights its flexibility and broader applicability. Besides, the results on large-scale PCTSP-500 and PCTSP-1000 in Appendix~\ref{app:exp}, Table~\ref{tab:large} demonstrate the scalability of DIFU-Ada. 

\begin{table*}[t]\Large
\renewcommand{\arraystretch}{1.25}
\centering
\resizebox{1.0\textwidth}{!}{
    \begin{tabular}{llccccccccc}
    \hline
    \hline
     &  & \multicolumn{2}{c}{\textbf{PCTSP-20}} & \multicolumn{2}{c}{\textbf{PCTSP-50}} & \multicolumn{2}{c}{\textbf{PCTSP-100}} & & & \\
    \cmidrule(l){3-4}
    \cmidrule(l){5-6}
    \cmidrule(l){7-8}
     & Method & \textbf{Gap} $\downarrow$ & \textbf{Time} $\downarrow$ & \textbf{Gap} $\downarrow$ & \textbf{Time} $\downarrow$ & \textbf{Gap} $\downarrow$ & \textbf{Time} $\downarrow$ & \textbf{Avg Gap} $\downarrow$ & \textbf{Training-free} & \textbf{Training Time} $\downarrow$ \\
    \hline
    \multirow{4}{*}{\rotatebox[origin=c]{90}{OR}} 
    & Gurobi & \textbf{\underline{0.00\%}} & 3.10s & --- & --- & --- & --- & --- & --- & --- \\
    & OR-Tools & 2.13\% & 12.31s  & 4.85\% & 2.02m  & 10.33\% & 5.84m & 5.77\% & --- & --- \\
    & ILS (C++)  & 1.07\% & 2.13s  & \textbf{\underline{0.00\%}} & 18.30s  & \textbf{\underline{0.00\%}} & 56.11s & 0.36\% & --- & --- \\
    & ILS (Python 10x)$^*$   & 63.23\% & 3.05s  & 148.05\% & 4.70s  & 209.78\% & 5.27s & 140.35\% & --- & --- \\
    \hline
    \multirow{8}{*}{\rotatebox[origin=c]{90}{Learning}}
     & AM (Greedy)  & 2.76\% & 0.02s  & 18.20\% & 0.07s  & 28.98\% & 0.15s & 16.65\% & {\ding{55}} & 3.5 days \\
     & AM (Sampling)  & 2.54\% & 2.43s & 14.58\% & 7.08s & 22.20\% & 15.13s & 13.11\% & {\ding{55}} & 3.5 days \\
     & MDAM$^*$ (Greedy) & 11.76\% & 41.10s  & 24.73\% & 1.31m & 30.07\% & 1.96m & 22.19\% & {\ding{55}} & 4.3 days \\
     & MDAM$^*$ (Beam Search)  & 5.88\% & 2.70m  & 18.81\% & 4.77m & 26.09\% & 6.97m & 16.93\% & {\ding{55}} & 4.3 days \\
     & ASP$^*$  & 12.05\% & 0.03s  & 10.34\% & 0.08s & \textbf{\underline{11.56\%}} & 0.18s & 11.32\% & {\ding{55}} & 4.6 days \\
     & AM-FT (greedy) & 2.11\% & 0.03s & 16.58\% & 0.07s & 29.08\% & 0.16s & 15.92\% & {\ding{55}} & 4.9 days \\
    & AM-FT (Sampling) & \textbf{\underline{1.02\%}} & 2.51s & 14.11\% & 8.02s & 25.19\% & 17.21s & 13.44\% & {\ding{55}} & 4.9 days \\
    
    & \textbf{DIFU-Ada (Ours)} & 4.45\% & 1.80s & \textbf{\underline{8.43\%}} & 1.92s & 16.22\% & 2.87s & \textbf{\underline{9.70\%}} & {\ding{51}} & \textbf{0 day} \\
    \hline
    \hline
    \end{tabular}
}
\caption{Comprehensive evaluation of cross-scale generalization capabilities across different solver categories on PCTSP instances. Comparison includes exact solvers (Gurobi), OR-based heuristics (OR-Tools, ILS), learning-based models (AM, MDAM, ASP, AM-FT, DIFU-Ada). Performance metrics include optimality gap, inference time, and training time. The results marked with $*$ are reported from \cite{wang2024asp}. The proposed DIFU-Ada achieves competitive performance while requiring no training.}
\label{tab:pctsp}
\end{table*}

\begin{table*}[t]\Large
\renewcommand{\arraystretch}{1.25}
\centering
\resizebox{1.0\textwidth}{!}{
\begin{tabular}{llccccccccc}
\hline
\hline
 &  & \multicolumn{2}{c}{\textbf{OP-20}} & \multicolumn{2}{c}{\textbf{OP-50}} & \multicolumn{2}{c}{\textbf{OP-100}} &  & & \\
\cmidrule(l){3-4}
\cmidrule(l){5-6}
\cmidrule(l){7-8}
 & Method & \textbf{Gap} $\downarrow$ & \textbf{Time} $\downarrow$ & \textbf{Gap} $\downarrow$ & \textbf{Time} $\downarrow$ & \textbf{Gap} $\downarrow$ & \textbf{Time} $\downarrow$ & \textbf{Avg Gap} $\downarrow$ & \textbf{Training-free} & \textbf{Training Time} $\downarrow$ \\
\hline
\multirow{4}{*}{\rotatebox[origin=c]{90}{OR}} 
& Gurobi & \textbf{\underline{0.00\%}} & 10.22s & --- & --- & --- & --- & --- & --- & --- \\
& Compass & 0.15\% & 0.27s & \textbf{\underline{0.00\%}} & 1.02s & \textbf{\underline{0.00\%}} & 6.74s & 0.05\% & --- & --- \\
& Tsili (Greedy)$^*$ & 16.58\% & 0.02s & 19.22\% & 0.03s & 19.71\% & 0.03s & 18.50\% & --- & --- \\
& Tsili (Sampling)$^*$ & 0.85\% & 0.55s & 4.46\% & 2.45s & 8.56\% & 9.08s & 4.62\% & --- & --- \\
\hline
\multirow{5}{*}{\rotatebox[origin=c]{90}{Learning}}
& AM (Greedy) & 5.78\% & 0.04s & 12.10\% & 0.09s & 28.75\% & 0.19s & 15.54\% & {\ding{55}} & 3.2 days \\
& AM (Sampling) & 2.62\% & 2.87s & 8.81\% & 8.12s & 22.17\% & 14.98s & 11.20\% & {\ding{55}} & 3.2 days \\
& AM-FT (Greedy) & 5.50\% & 0.04s & 11.51\% & 0.08s & 23.40\% & 0.18s & 13.47\% & {\ding{55}} & 4.7 days \\
& AM-FT (Sampling) & \textbf{\underline{1.01\%}} & 2.76s & 7.19\% & 8.44s & 19.51\% & 18.15s & 9.24\% & {\ding{55}} & 4.7 days \\

& \textbf{DIFU-Ada (Ours)} & 3.44\% & 2.01s & \textbf{\underline{6.11\%}} & 2.29s & \textbf{\underline{12.21\%}} & 2.85s & \textbf{\underline{7.26\%}} & {\ding{51}} & \textbf{0 day} \\
\hline
\hline
\end{tabular}
}
\caption{Comprehensive evaluation of cross-scale generalization capabilities across different solver categories on OP instances. Comparison includes exact solvers (Gurobi), OR-based heuristics (Compass, Tsili), learning-based models (AM, AM-FT, DIFU-Ada). The results marked with $*$ are reported from \cite{lin2024cross}.}
\label{tab:op}
\end{table*}

\begin{figure}[b]
    \centering
    \includegraphics[width=0.99\linewidth]{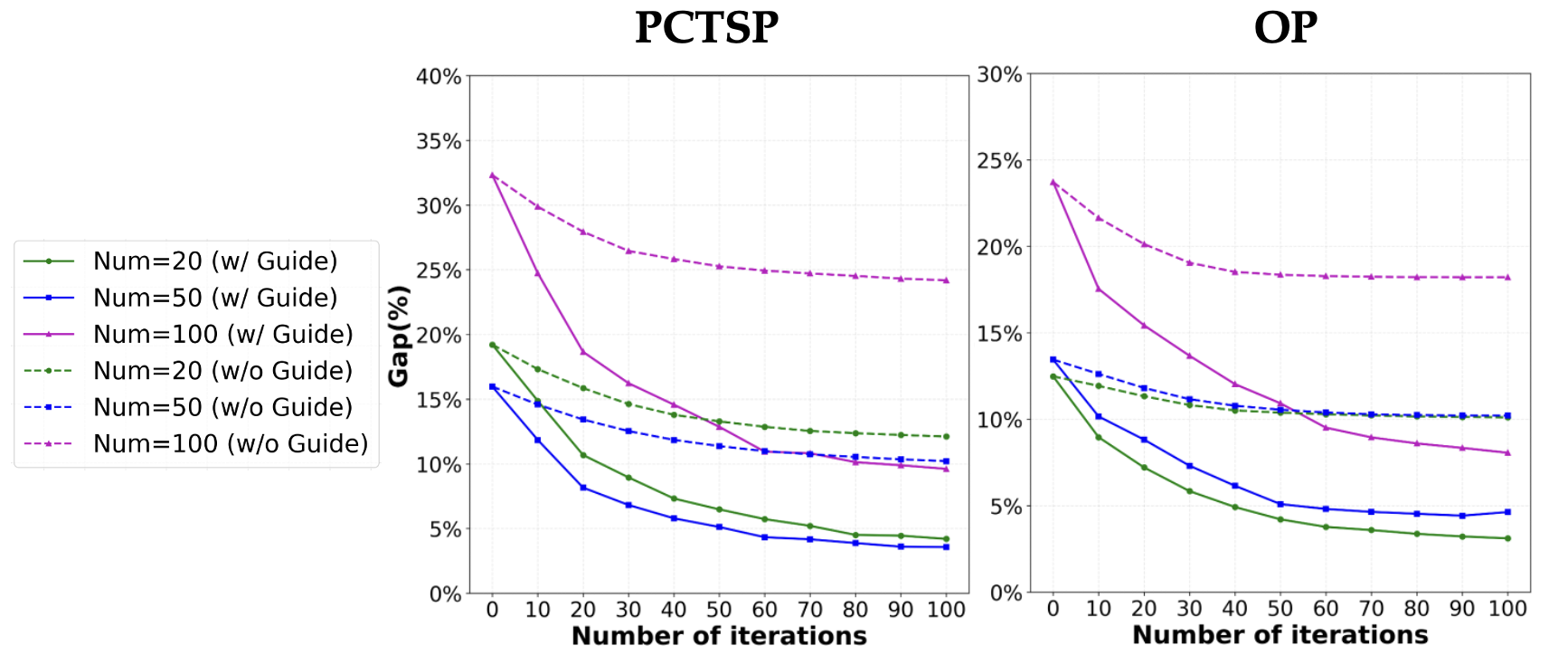}
    \caption{The Optimality Gap of performing recursive travel with or without energy-guided sampling on PCTSP and OP.}
    \label{img:abl_1}
\end{figure}

\subsection{Ablation Studies}

\begin{figure}[htbp] 
\centering    
\subfigure{  
\includegraphics[width=0.45\columnwidth]{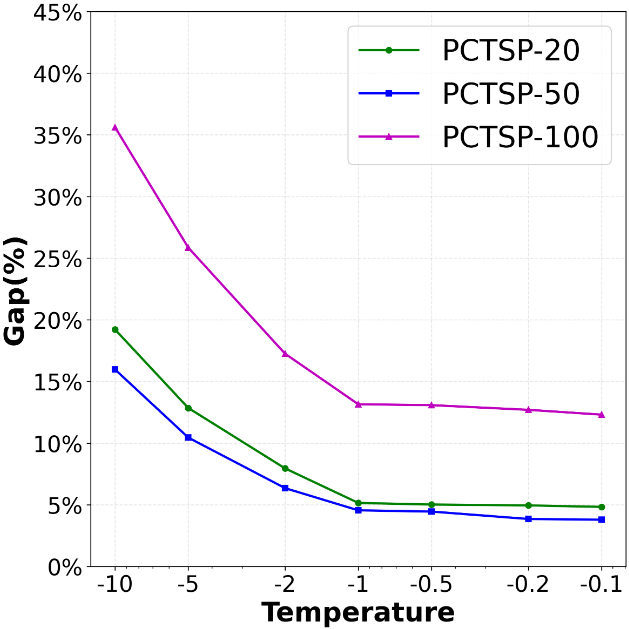}
}
\subfigure{     
\includegraphics[width=0.45\columnwidth]{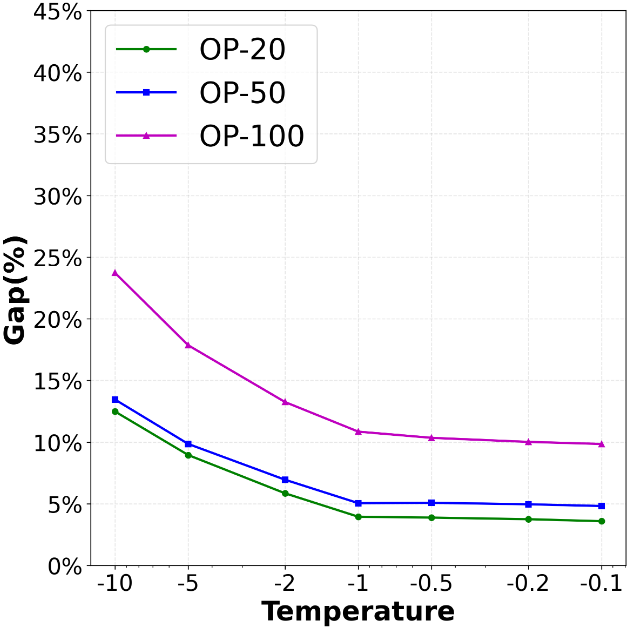}     
}
\caption{Optimality gap changes with respect to the guided temperature ($-\lambda$ as the x-axis label) on PCTSP and OP.}
\label{Fig: temp}       
\end{figure}

The DIFU-Ada framework integrates two key components: energy-guided sampling and recursive renoising-denosing travel. To assess the contribution of each, we conducted ablation studies. Figure~\ref{Fig: tradeoff} quantitatively illustrates the impact of the recursive renoising-denosing, demonstrating a progressive enhancement in cross-problem transfer capabilities as the number of recursive iterations increases. Figure~\ref{img:abl_1} presents a analysis of performance with or without energy-guided sampling during the recursive iterations. The performance gap observed in these studies show the efficacy of both components and highlight their distinct roles in the overall effectiveness to achieve zero-shot cross-problem transfer.

Moreover, Figure~\ref{Fig: temp} shows the effect of the temperature parameter in controlling the guidance strength of adaptation. Higher temperature effectively preserve the structural information in the original problem instances, whereas lower temperatures adapt to the characteristics of new problem instances. This temperature-dependent behavior enables fine-grained control over the balance between maintaining prior knowledge and new problem features. In Appendix~\ref{app:exp}, Figure~\ref{fig:mu}, the results on varying hyperparameter $\mu$ in energy function show that the performances are not sensitive to the choice of $\mu$, thus we choose the optimal $\mu=1$ empirically for all problems.

\section{Conclusion}

We introduce DIFU-Ada, an inference-time adaptation framework for zero-shot cross-problem transfer in diffusion-based solvers. By integrating energy-guided sampling with a recursive denoising strategy, our approach adapts pre-trained models to generate high-quality solutions for new TSP variants without retraining. Theoretical analysis supports these transfer capabilities, and experiments on PCTSP and OP confirm competitive zero-shot performance against existing learning-based methods.

This work highlights the potential of adapting diffusion solvers for real-world combinatorial optimization problems with dynamic constraints. Our framework offers a flexible, off-the-shelf sampling scheme that avoids costly retraining while enhancing cross-problem and cross-scale generalization. While promising on TSP variants, a key limitation is its validation on routing problems; future work should therefore test its applicability across a broader spectrum of combinatorial domains.

\bibliographystyle{unsrt}
\bibliography{main}

\begin{appendices}
\section{Problem Descriptions of TSP Variants}
\label{app:def}
\begin{itemize}
\item \textbf{Traveling Salesman Problem (TSP)} requires finding the minimal-length Hamiltonian cycle in a complete graph, where the salesman must visit each city exactly once before returning to the starting point. \\
\item \textbf{Prize Collecting TSP (PCTSP)} \cite{balas1989prize} extends the classical TSP by introducing node-specific prizes and penalties. The objective is to optimize a trade-off between minimizing tour length and unvisited node penalties while ensuring collected prizes exceed a predefined threshold. This formulation creates a more complex optimization landscape where node visitation decisions must balance multiple competing factors. \\
\item \textbf{The Orienteering Problem (OP)}, first introduced by \cite{golden1987orienteering}, is a fundamental combinatorial optimization problem with widespread applications in real-world scenarios. In the OP, each node in the network is associated with a non-negative prize value, and the objective is to determine an optimal tour that begins and ends at a designated depot node. The tour must satisfy two key constraints: maximize the total collected prizes from the visited nodes, and ensure the total tour length does not exceed a predetermined distance limit. This problem effectively captures the trade-off between reward collection and resource constraints.
\end{itemize}

\section{Experimental Setup Details} \label{setup}

\textbf{Dataset.} We evaluate our approach on two common TSP variants: the Prize Collecting Traveling Salesman Problem (PCTSP) and the Orienteering Problem (OP). Since our method operates in a zero-shot setting, no training or fine-tuning datasets for PCTSP or OP were used. For evaluation, we generated test instances following the standard procedure from prior work \cite{kool2018attention, lin2024cross}. Specifically, for each problem scale $N$, we randomly generated the 2D coordinates of $N$ nodes within a $(0,1)^N$ space. For PCTSP, the depot is the first node, and the maximum tour length is set to $N/2$. For OP, prizes are sampled uniformly from $(0,1)$. The reference solutions for evaluation were generated by the exact solver Gurobi for 20-node instances. For larger 50- and 100-node instances where exact solving is intractable, we used high-performance heuristic solvers: ILS for PCTSP and Compass for OP. \\

\noindent\textbf{Evaluation.} We follow the standard evaluation protocol from \cite{kool2018attention}, generating a test set of 10,000 instances for each problem scale ($N \in {20, 50, 100}$). Model performance is assessed using two primary metrics: the average solution cost (or tour length) and the optimality gap, calculated as $(\text{Cost model} / \text{Cost baseline} - 1) * 100\%$. For computational efficiency, we measure the per-instance inference time. While we also consider the training time of baselines, it is a key advantage of our method that it requires zero training or fine-tuning on the target problems. \\

\noindent\textbf{Baselines.} We compare DIFU-Ada against three categories of baselines. For each learning-based method, we clarify its training paradigm to ensure a fair comparison:

\begin{itemize}
    \item Exact/Heuristic Solvers: For PCTSP, we use Gurobi and Iterated Local Search (ILS). For OP, we use Gurobi, Compass \cite{kobeaga2018efficient}, and Tsili \cite{tsiligirides1984heuristic}.
    \item Learning-based (Trained/Fine-tuned): These methods are either trained from scratch or fine-tuned on the target problem (PCTSP or OP). This category includes AM \cite{kool2018attention}, MDAM \cite{xin2021multi}, ASP \cite{wang2024asp} and AM-FT \cite{lin2024cross}.
    \item Learning-based (Zero-shot): These methods, like ours, are pre-trained on TSP and applied directly to the target problems without further training. This category includes the original DIFUSCO \cite{sun2023difusco}, and T2T \cite{li2024distribution}. However, these two methods are specifically optimized for solving TSP problems, our proposed DIFU-Ada offers the only zero-shot cross-problem ability without additional training.
\end{itemize}

\noindent\textbf{Experimental Setup.} The proposed Inference Time Adaptation for Diffusion-based solvers (\textbf{DIFU-Ada}) builds upon DIFUSCO's \cite{sun2023difusco} TSP-trained checkpoints in three different scales without additional training.

The backbone architecture used in diffusion model is the same as DIFUSCO, building upon a GNN Encoder architecture which, for our experiments, is configured with 12 GNN layers and a hidden dimension of 128 for all feature representations. The core of the model is the Gated Graph Convolutional Network. This layer updates both node and edge features by applying a series of linear transformations followed by a sigmoid gating mechanism, with sum aggregation and Layer Normalization used by default. For input processing, the model first converts raw node coordinates into 128-dimensional features using a PositionEmbeddingSine module. The diffusion timestep t is also embedded into a vector representation and injected into each of the 12 GNN layers, allowing the model to condition its predictions on the noise level. Residual connections are employed around each GNN layer to ensure stable training. For computational efficiency, our implementation operates on sparse graph representations.

The results for DIFU-Ada are recorded over 100 recursive iterations, a value found to provide a strong balance between solution quality and inference time. Each iteration consists of a 5-step re-noising phase followed by a 1-step guided denoising phase. We employ a greedy decoding strategy for final solution construction. The energy functions guiding the sampling are defined in Eq.~\ref{eq.pctsp} and \ref{eq.op}. To demonstrate robustness without problem-specific tuning, the guidance temperature $\lambda$ was fixed to 0.1 and the constraint coefficient $\mu$ was set to 1 for all instances. All our experiments were conducted on a single NVIDIA Tesla V100 GPU using PyTorch 2.1. Baseline results are taken from their respective publications, and any direct comparison of inference times should consider potential differences in hardware.\\

\section{Theoretical Analysis of Transferring Knowledge from Pre-trained Model} \label{Appendix:analysis}
Consider the scenario where we aim to transfer knowledge from the original Traveling Salesman Problem (TSP) to its variants: the Prize Collecting TSP (PCTSP) and the Orienteering Problem (OP). The diffusion model, trained on optimal TSP instances, estimates $p_\theta(\mathbf{x}_t|G')$ such that, under the assumption of perfect optimization, $p_\theta (\mathbf{x}_0 | G')$ approximates $q_{\textup{TSP}}(\mathbf{x}_0 | G')$, where the latter represents the distribution of ground-truth TSP solutions for instance $G'$. Through the Bayesian equation established in \eqref{eq:log_decomp}, we expect such a pre-trained prior would substantially enhance posterior sampling solution quality, as TSP shares fundamental structural similarities with PCTSP and OP, thereby encoding relevant domain knowledge. We formalize such similarities in the following analysis. We make use of the following definition to facilitate the analysis. 

\begin{definition}[Marginal Decrease]
\label{def:md_main}
   For a non-empty subset of nodes $S \subseteq V$, let $\textup{TSP}(S)$ denote the cost of the optimal TSP tour visiting all nodes in $S$. The marginal decrease of a subset $S$ is defined as 
   \[
   \Delta(S) = \textup{TSP}(V) - \textup{TSP}(V\setminus S).
   \]
\end{definition}

The marginal decrease measures the cost reduction of not visiting a subset of nodes $S$, which helps quantify the difference between the optimal tours. Take PCTSP as an example. If for any non-empty subset of nodes $S \subseteq V$, the penalty of not visiting the nodes in $S$ satisfies 
$\sum_{i\in S} {p_i} \geq \Delta(S)$,
then PCTSP and TSP share the same optimal tours. Based on this notion, we formalize the structural similarities in the following theorem.

\begin{theorem}\label{thm:prior_bound}
     For a non-empty subset of nodes $S\subseteq V$, let $\textup{TSP}(S)$ and $\textup{argTSP}(S)$ denote the optimal cost and optimal tours of TSP on the subgraph specified by $S$. Under Assumptions\footnote{We assume simpler setups for PCTSP and OP for clearer presentation of the underlying similarities. These assumptions can be relaxed as discussed in Appendix \ref{app:thm-ext}.} \ref{ass:pctsp} and \ref{ass:op}, the optimal tours of PCTSP are
     $\textup{argTSP}(V\setminus S_{\textup{PCTSP}})$, where
     \[
     S_{\textup{PCTSP}} \in \arg\min_{S\subseteq V} {\sum_{i\in S} {p_i} - \Delta(S)},
     \]
     and the optimal tours of OP are $\textup{argTSP}(V\setminus S_{\textup{OP}})$, where
    \[
    \begin{aligned}
    S_{\textup{OP}} \in &\arg\min_{S\subseteq V} {\Delta(S)} \\
    &\qquad\textup{ s.t. } \Delta(S) \geq \textup{TSP}(V) -D_{\textup{OP}}.
    \end{aligned}
    \]
    Here $D_{\textup{OP}}$ is the distance limit  of the total tour length, and $\Delta(S)$ denotes the marginal decrease of a non-empty subset of nodes $S$. 
\end{theorem}

\noindent\textbf{Proof.} To analyze the relation between the optimal solutions of TSP and that of PCTSP/OP, we make use of the following definitions. 

\begin{definition}[Marginal Decrease]
\label{def:md}
   For a non-empty subset of nodes $S \subseteq V$, let $\textup{TSP}(S)$ denote the cost of the optimal TSP tour visiting all nodes in $S$. The marginal decrease of a subset $S$ is defined as 
   \[
   \Delta(S) = \textup{TSP}(V) - \textup{TSP}(V\setminus S).
   \]
\end{definition}

The marginal decrease measures the cost reduction of not visiting a subset of nodes $S$. This concept helps quantify the relation between the optimal solutions of TSP/PCTSP/OP. We first make the following assumptions on the PCTSP and OP problems in the analysis. Note that we make these assumptions just for the simplicity of presentation.

\begin{assumption} [PCTSP Setting]
\label{ass:pctsp}
     We consider PCTSP without the minimum prize constraint.
\end{assumption}

\begin{assumption}[OP Setting]
\label{ass:op}
    For OP, we assume an identical reward at each city.
\end{assumption}

Note that the above assumptions can be relaxed by considering a node-weighted TSP, which can be transformed into a standard TSP as discussed in Appendix \ref{app:thm-ext}. 


Under these assumptions, we compare the optimal solutions of PCTSP/OP with that of TSP as follows.

(i) For PCTSP, if for any non-empty subset of nodes $S \subseteq V$, the penalty of not visiting the nodes in $S$ satisfies 
\[
\sum_{i\in S} {p_i} \geq \Delta(S), 
\]
then PCTSP and TSP share the same optimal solutions, since not visiting any subset of nodes would result in a higher total cost. We can thus formulate the optimal cost (tour length + penalty) of PCTSP as
\begin{equation}
\label{eq:pctsp_opt}
    \text{PCTSP}(V) = \text{TSP}(V) + \min_{S\subseteq V} {\sum_{i\in S} {p_i} - \Delta(S)}.
\end{equation}
And we have the optimal solutions of PCTSP being the tours in $\text{TSP}(V\setminus S)$ with 
\[
S \in \arg\min_{S\subseteq V} {\sum_{i\in S} {p_i} - \Delta(S)},
\]
completing the proof for PCTSP.

(ii) For OP, since we assume a uniform reward at each city, the optimization objective becomes visiting as much nodes as possible under the distance limit of the total tour length. This objective is identical to minimizing the travel cost while respecting the budget, because the optimal cost of TSP satisfies the following property:
\[
    \text{For any non-empty subsets of nodes } S, V, \text{ if }  S\subseteq V, \text{ then } \text{TSP}(S) \leq \text{TSP}(V). 
\]
We can thus formulate the optimal tour length of OP as
\begin{align}
 \text{OP}(V) ={}& \max_{S\subseteq V} {\text{TSP}(V\setminus S)} \nonumber\\ &\ \ \text{ s.t. } \text{TSP}(V\setminus S) \leq D_{\text{OP}} \nonumber \\
 ={}& \text{TSP}(V) + \min_{S\subseteq V} {\Delta(S)} \label{eq:op_opt} \\ & \qquad\qquad \ \ \ \  \text{ s.t. } \Delta(S) \geq \text{TSP}(V) -D_{\text{OP}}, \nonumber
\end{align}
where $D_{\text{OP}}$ is the distance limit of OP. And we have the optimal solutions of OP being the tours in $\text{TSP}(V\setminus S)$ with 
\[
\begin{aligned}
S \in &\arg\min_{S\subseteq V} {\Delta(S)} \\
&\qquad\text{ s.t. } \Delta(S) \geq \text{TSP}(V) -D_{\text{OP}},
\end{aligned}
\]
completing the proof for OP.

Theorem \ref{thm:prior_bound} shows that the optimal solutions of PCTSP and OP are indeed the optimal tours of TSP on some subgraph of $G'$, revealing their fundamental structural similarities. We expect that the pre-trained diffusion model is powerful enough to generate high-quality TSP solutions given test instances with various graph sizes. We can therefore understand the inference time adaptation in this scenario as forcing the pre-trained model to focus on a subgraph of $G'$. Through this theoretical analysis, we deepen our understanding of how the proposed training-free inference time adaptation works when transferring from a pre-trained network to its variants. Although we focus on PCTSP and OP in this paper, we show that our methodology and insight also hold for more challenging TSP variants such as TSP with time windows in Appendix \ref{app:tsp-tw}. 

\section{Relaxing Assumptions~\ref{ass:pctsp} and~\ref{ass:op}}
\label{app:thm-ext}

Our analysis in Theorem \ref{thm:prior_bound} can be extended to the cases with minimum prize constraints or non-uniform rewards by considering node-weighted TSP formulations. The key is that the optimal solutions of PCTSP/OP are the optimal solutions of certain node-weighted TSP over a subgraph as shown below.

(i) For the general case of OP, we can reformulate the problem by assigning each node $i$ a score $s_i$ equal to its negative reward and setting all edge costs to zero. Meanwhile, we maintain the edge cost function $w_e$ solely to represent the budget constraint of OP. This transforms OP into a node-weighted TSP with an additional budget constraint. 

(ii) Similarly, for PCTSP, we assign $s_i = -p_i$ (the negative penalty of node $i$) while preserving the original edge costs. The optimal solution to PCTSP then corresponds to that of a node-weighted TSP with minimum prize constraints.

Since the pre-trained model was trained on standard TSP instances, we can convert node-weighted TSP to a standard one by splitting each node $i$ into two nodes $i_{in}$ and $i_{out}$, and adding a directed edge from $i_{in}$ to $i_{out}$ with weight $s_i$. Given this conversion, we could explain the effectiveness of our method as follows: the energy-guided sampling implicitly considers the node-weighted TSP and guides the generation to zoom in a subgraph of the node-weighted graph which respects the additional constraints.

\section{Adaptation to TSP with Time Windows (TSP-TW)}
\label{app:tsp-tw}

TSP with Time Windows (TSP-TW) is a variant of TSP that requires each node $i$ to be visited within a given time range $[e_i, l_i]$. Our approach can also handle TSP-TW through a time-expanded graph transformation:

For each node $i$ with time window $[e_i, l_i]$, we create multiple time-indexed replicas $(i_{e_i}, i_{e_i+1}, ..., i_{l_i})$. The original graph's edges are then transferred to this expanded representation according to temporal feasibility - if node $i$ connects to node $j$ in the original graph, we connect $i_t$ to $j_{t+1}$ in the expanded graph (assuming unit travel time for simplicity).

This transformation converts the time window constraints into a structural pattern within the expanded graph. The TSP-TW solution corresponds to a standard TSP tour on this expanded graph with one critical constraint: exactly one replica of each original node must be visited, expressed as:
\[
\sum_{t\in[e_i,l_i]} \sum_{(j,i_t)\in E} x_{j,i_t} = 1.
\]
This constraint can be directly encoded as an energy function in our framework. Our method then proceeds by:
\begin{itemize}
    \item Sampling initial TSP solutions on the time-expanded graph
    \item Using energy-guided sampling to enforce the single-visit-per-node constraint
    \item Converging toward feasible TSP-TW solutions
\end{itemize}

Through this transformation, TSP-TW effectively becomes a "subgraph TSP" scenario similar to that of PCTSP/OP, demonstrating the flexibility of our energy-guided approach for handling complex constraints without requiring problem-specific training.

\section{Additional Experimental Results} \label{app:exp}

In this part, we provide several additional experimental results. 

Table~\ref{tab:large} underscores the scalability of our approach on large-scale PCTSP-500 and PCTSP-1000 problems. While heuristic solvers like OR-Tools are prohibitively slow and other learning-based methods require dedicated training to achieve top performance, our training-free DIFU-Ada remains highly competitive. It achieves solution costs (14.5 and 21.7) that are remarkably close to the state-of-the-art trained model, GLOP-S (14.3 and 19.8), while entirely circumventing the need for problem-specific training or fine-tuning. These results validate DIFU-Ada as a robust and efficient framework that balances high performance with zero training overhead, even at a large scale.

\begin{table}[h!]
\centering
\begin{tabular}{lccccc}
\toprule
 & \multicolumn{2}{c}{\textbf{PCTSP-500}} & \multicolumn{2}{c}{\textbf{PCTSP-1000}} & \\
\cmidrule(lr){2-3} \cmidrule(lr){4-5}
Method & \textbf{Cost$\downarrow$} & \textbf{Time$\downarrow$}  & \textbf{Cost$\downarrow$}  & \textbf{Time$\downarrow$}   & \textbf{Training-free}  \\
\midrule
OR Tools* & 15.0 & 1h & 24.9 & 1h & -- \\
OR Tools (more iterations)* & 14.4 & 16h & 20.6 & 16h & -- \\
\midrule
AM* & 19.3 & 14m & 34.8 & 23m & \textcolor{red}{\ding{55}} \\
MDAM* & 14.8 & 2.8m & 22.2 & 17m & \textcolor{red}{\ding{55}} \\
GLOP-S* & \textbf{14.3} & 1.5m & \textbf{19.8} & 2.5m & \textcolor{red}{\ding{55}} \\
\midrule
\textbf{DIFU-Ada (Ours)} & 14.5 & 6.6m & 21.7 & 8.1m & \textcolor{green}{\ding{51}} \\
\bottomrule
\end{tabular}
\caption{Comparison of various methods on 128 instances of PCTSP-500 and PCTSP-1000. The data generation and evaluations are conducted following Ye et al. (2024). Results marked with * are taken directly from Ye et al. (2024).}
\label{tab:large}
\end{table}

Table~\ref{tab:results} provides a focused evaluation of leading diffusion-based CO solvers on the task of cross-scale generalization for TSP. This comparison demonstrates a key strength of our method: although DIFU-Ada is designed for the more complex challenge of both cross-problem and cross-scale generalization, its performance on the standalone cross-scale TSP task remains highly competitive. The results show that DIFU-Ada achieves performance comparable to the current state-of-the-art, FastT2T, confirming that its enhanced capabilities do not compromise its effectiveness on foundational benchmarks TSP.\\

\begin{table*}[h]
\centering
\renewcommand{\arraystretch}{1.2} 
\setlength{\tabcolsep}{6pt} 
\begin{tabular}{llccc}
\toprule
\multirow{2}{*}{\textbf{Training}} & \multirow{2}{*}{\textbf{Method}} & \multicolumn{3}{c}{\textbf{Testing Dataset}} \\ 
\cmidrule(lr){3-5}
 &  & \textbf{TSP-20} & \textbf{TSP-50} & \textbf{TSP-100} \\
\midrule
\multirow{4}{*}{\textbf{TSP-20}} 
 & DIFUSCO & 3.14, 0.23\% & 5.75, 0.43\% & 7.90, 1.81\% \\
 & T2T & 3.14, 0.12\% & 5.71, 0.19\% & 7.82, 0.87\% \\
 & Fast T2T & \textbf{3.12, 0.03\%} & \textbf{5.69, 0.00\%} & \textbf{7.76, 0.08\%} \\
 & \textbf{Ours} & 3.14, 0.15\% & 5.71, 0.22\% & 7.83, 0.95\% \\
\midrule
\multirow{4}{*}{\textbf{TSP-50}} 
 & DIFUSCO & 3.14, 0.25\% & 5.69, 0.09\% & 7.87, 1.44\% \\
 & T2T  & 3.14, 0.11\% & 5.69, 0.02\% & 7.80, 0.55\% \\
 & Fast T2T & \textbf{3.12, 0.05\%}  & \textbf{5.69, 0.00\%} & \textbf{7.76, 0.08\%}\\
 & \textbf{Ours} & 3.14, 0.14\% & 5.69, 0.05\% & 7.81, 0.60\%\\
\midrule
\multirow{4}{*}{\textbf{TSP-100}} 
 & DIFUSCO & 3.16, 0.44\% & 5.70, 0.25\% & 7.78, 0.23\% \\
 & T2T & 3.14, 0.21\% & 5.70, 0.11\% & 7.77, 0.08\% \\
 & Fast T2T & \textbf{3.12, 0.10\%} & \textbf{5.69, 0.01\%} & \textbf{7.76, 0.01\%} \\
 & \textbf{Ours} & 3.14, 0.20\% & 5.71, 0.25\% & 7.77, 0.09\% \\
\bottomrule
\end{tabular}
\caption{Results of various diffusion-based methods \cite{sun2023difusco, li2024distribution, li2024fast} on different TSP testing datasets (TSP-20, TSP-50, TSP-100), only considering cross-scale generalization capability.}
\label{tab:results}
\end{table*}

\begin{figure}[h!]
    \centering
    \includegraphics[width=0.8\linewidth]{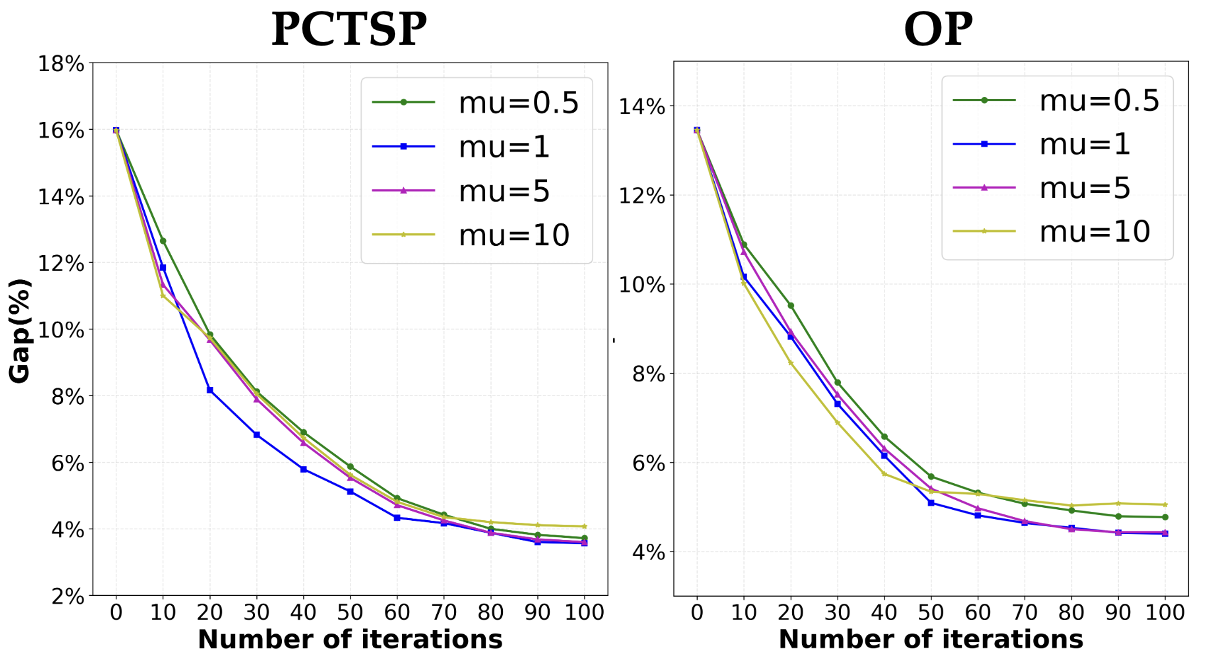}
    \caption{Impact of the energy function hyperparameter $\mu$ on the optimality gap for PCTSP-50 and OP-50.}
    \label{fig:mu}
\end{figure}

Figure~\ref{fig:mu} illustrates the performance sensitivity of our method to the constraint coefficient $\mu$ for PCTSP-50 and OP-50. The results indicate that while the choice of $\mu$ can slightly alter the convergence speed, the final solution quality remains remarkably stable across a range of values. For both problem types, we observe only a minor performance gap between the different settings. This analysis confirms that our method is robust to the specific choice of $\mu$. Therefore, to simplify the experimental setup and highlight the generalizability of our approach, we use a fixed $\mu=1$ for all experiments, avoiding the need for problem-specific hyperparameter tuning. \\

\begin{figure}[h!]
    \centering
    \includegraphics[width=0.95\linewidth]{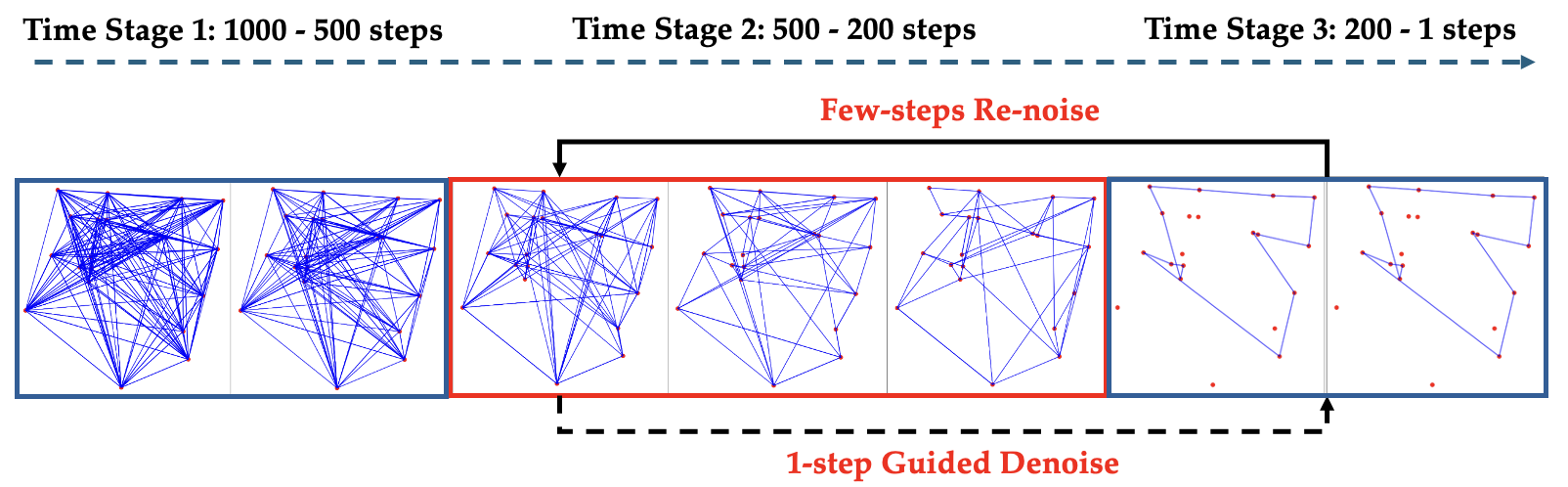}
    \caption{Visualized results of applying recursive renoising-denoising to gradually modify the noisy heatmap on PCTSP-20.}
    \label{fig:visual}
\end{figure}

Figure~\ref{fig:visual} provides a visualization of the reverse process for a PCTSP instance. It depicts the evolution of the solution's adjacency matrix, represented as a heatmap, throughout our recursive renoising-denoising procedure. The process starts from a matrix corresponding to pure uniform noise and progressively denoises it into a structured matrix that represents a feasible PCTSP tour.

\end{appendices}
\end{document}